\documentclass{article}
\usepackage{ijcai21}

\usepackage{times}
\usepackage{soul}
\usepackage{url}
\usepackage[hidelinks]{hyperref}
\usepackage[utf8]{inputenc}
\usepackage[small]{caption}
\usepackage{graphicx}
\usepackage{amsmath}
\usepackage{amsthm}
\usepackage{booktabs}
\usepackage[linesnumbered,ruled,vlined]{algorithm2e}
\usepackage{comment}
\usepackage{multirow}

\usepackage{amssymb,bm}

\newcommand{\featuretarget}{{\mathbf X}^{\mathit{tgt}}}
\newcommand{\featurestation}{{\mathbf X}^{\mathit{stn}}_s}
\newcommand{\featurecity}{{\mathbf X}^{\mathit{city}}}
\newcommand{\sourcecities}{C_{\mathit{src}}}

\newcommand{\targetcity}{c_{\mathit{tgt}}}
\newcommand{\sourcecity}{c_{k}}
\newcommand{\targetlocation}{l_{\mathit{tgt}}}

\newcommand{\airdata}{D^A}
\newcommand{\poidata}{D^P}
\newcommand{\roaddata}{D^R}
\newcommand{\meteodata}{D^M}

\newcommand{\losspred}{\mathcal{L}_{{\mathit f}}}
\newcommand{\lossmean}{\mathcal{L}_{{\mathit e}}}
\newcommand{\lossadv}{\mathcal{L}_{{\mathit a}}}

\newcommand{\Category}[1]{\Upsilon_{#1}}
\newcommand{\category}{\upsilon}

\theoremstyle{definition}
\newtheorem{definition}{Definition}

\makeatletter
\def\Hline{
  \noalign{\ifnum0=`}\fi\hrule \@height 3.\arrayrulewidth \futurelet
  \reserved@a\@xhline}
\makeatother

\title{AIREX: Neural Network-based Approach for Air Quality Inference in Unmonitored Cities}

\author{
Yuya Sasaki$^1$
\and
Kei Harada$^1$\and
Shohei Yamasaki$^{1}$\And
Makoto Onizuka$^1$
\affiliations
$^1$Osaka university\\
\emails
\{sasaki, harada.kei, yamasaki.shohei, onizuka\}@ist.osaka-u.ac.jp
}

\begin{document}

\maketitle


\begin{abstract}

Urban air pollution is a major environmental problem affecting human health and quality of life. 
Monitoring stations have been established to continuously obtain air quality information, but they do not cover all areas. 
Thus, there are numerous methods for spatially fine-grained air quality inference. 
Since existing methods aim to infer air quality of locations only in monitored cities, they do not assume inferring air quality in {\it unmonitored} cities. 

In this paper, we first study the air quality inference in unmonitored cities.
To accurately infer air quality in unmonitored cities, we propose a neural network-based approach {\it AIREX}. 
The novelty of AIREX is employing a mixture-of-experts approach, which is a machine learning technique based on the divide-and-conquer principle, to learn correlations of air quality between multiple cities. 
To further boost the performance, it employs attention mechanisms to compute impacts of air quality inference from the monitored cities to the locations in the unmonitored city.
We show, through experiments on a real-world air quality dataset, that AIREX achieves higher accuracy than state-of-the-art methods.
\end{abstract}


\section{Introduction}
Urban air pollution poses a severe and global problem.
The fine-grained assessment of urban air quality is crucial for both the governments and citizens to establish means to improve human health and quality of life.
Monitoring stations have been established in numerous cities to continuously obtain air quality information. 
However, due to high construction and management costs, monitoring stations are sparsely installed and concentrated only in areas of higher importance, such as cities with large populations.
As a result, it is essential to infer air quality in areas without monitoring stations.

The development of neural network techniques has accelerated a neural network-based approach for inferring air quality of target locations~\cite{Zheng2013U-Air:Data,Cheng2018AStations,chang2004air}.
This approach leverages available external data related to the air quality, such as point-of-interest and meteorology, to capture features of locations.
The existing methods aim to infer air quality of target locations only in monitored cities (i.e., cities with monitoring stations).
They do not capture the difference of features in cities, which causes the low accuracy of air quality inference.
Since not all cities have monitoring stations, we cannot accurately infer the air quality of target locations within the unmonitored cities.

\smallskip
\noindent
{\bf Problem definition and challenges}: We study a new problem, {\it air quality inference in unmonitored cities}, to globally solve the urban air pollution problem.
A straightforward approach for the problem is the use of existing models that are trained by air quality data of cities in the vicinity of the target unmonitored city.
However, even the state-of-the-art method ADAIN~\cite{Cheng2018AStations} deteriorates the inference accuracy in unmonitored cities, even when using air quality data of numerous monitored cities as training data (see Table~\ref{tab:beijing_dist} in experiments section).

Therefore, we need a new neural network architecture in this problem.
We face two challenges: (1) how to design a neural network architecture to capture the correlations of air quality between monitored and unmonitored cities and (2) how to train models without available air quality data of the unmonitored cities.
For the first challenge, since features of cities differ, architectures must capture their differences and reflect them in the inference of air quality.
It is difficult to select optimal monitoring stations for model training due to the absence of air quality data in unmonitored cities.
For the second challenge, since we do not have air quality data of the unmonitored city, architectures must be trained only by using air quality data of monitored cities and external data.
We cannot directly learn the correlations between monitored and unmonitored cities.

In summery, we require a new neural network architecture that (1) can automatically capture the correlation between monitored and unmonitred cities without selecting monitoring stations and (2) can be trained in an unsupervised manner. 

\noindent
{\bf Contributions}: We propose a novel neural network-based architecture {\it AIREX}.
AIREX automatically captures the correlations between monitored and unmonitored cities. 
The novel design of AIREX is based on the effective combination of the mixture-of-experts~\cite{jacobs1991adaptive,masoudnia2014mixture,Guo2018Multi-sourceExperts} and attention mechanisms~\cite{bahdanau2014neural}.
The mixture-of-experts approach is a machine learning technique based on the divide-and-conquer principle.
This approach uses multiple models (called {\it experts}) and aggregates outputs of experts for deriving the final output.
Each expert in AIREX corresponds to individual monitored cities, and thus AIREX infers air quality in unmonitored cities by aggregating air quality assessed from individual monitored cities.
The attention mechanism further boosts the performance of AIREX.
AIREX employs two attentions to capture the importance of monitored cities and monitoring stations individually for computing weights of influences from monitored cities to the target location.
The effective combination of mixture-of-experts approach and attention mechanism achieves accurate air quality inference.

For training AIREX, we develop a training method using a meta-training approach~\cite{Guo2018Multi-sourceExperts}, which is suitable for training of the mixture-of-experts approach in an unsupervised manner.
In our training method, we regard one of the monitored cities as an unmonitored city at the training phase so that AIREX can be learned in an unsupervised manner.
We use multi-task learning~\cite{caruana1997multitask} for training both the whole AIREX and experts with capturing the difference among cities. 
This training method enables to learn the correlations between monitored and unmonitored cities without air quality data of unmonitored cities.

Our contributions presented in this study are as follows: 
\begin{itemize}
    \item We address a novel problem that infers air quality information in unmonitored cities by using the air quality data obtained from other cities. We show that state-of-the-art methods are not suitable for this problem.
    \item We propose AIREX that can accurately infer air quality information in unmonitored cities. This employs the mixture-of-experts approach and attention mechanism to capture the correlations of air quality between monitored and unmonitored cities.
    \item Through experiments with 20 cities in China, we show that AIREX achieves higher accuracy than the-state-of-the-art method. 
\end{itemize}


\section{Problem Formulation}


We describe the notations and definitions used in the formulation of the problem that we solve in this study.

There are two types of cities, namely, target and source cities, that denote unmonitored and monitored cities, respectively.
Each city $c$ has its representative specific location $l_c$ (e.g., the center of $c$).
We assume that we have a single target city $\targetcity$ and a set $\sourcecities$ of source cities.
We denote the set of monitoring stations by $S$ and each monitoring station $s \in S$ has its location $l_s$, which periodically monitors a quantity of air pollutants, such as PM$_{2.5}$, over the time domain $T=\langle t_1, t_2, \ldots, t_{|T|}\rangle$.
Source city $\sourcecity \in \sourcecities$ has a set of monitoring stations $S_k \subseteq S$.
We denote $s_{k,i}$ as monitoring station $s_i \in S_k$.
We define air pollutant data as follows:

\begin{definition}[Air pollutant data]
Air pollutant data $\airdata$ consists of quantities of air pollutants monitored by stations, and they are time-dependent.
\end{definition}



Cities have characteristics that affect air quality.
To infer air quality, we use three external data that were frequently employed in prior studies \cite{Xu2016WhenInference,Zheng2013U-Air:Data,Cheng2018AStations}; Point-of-interest (PoI), road network, and meteorological information.

\begin{definition}[PoI data]
PoI data $\poidata$ consist of PoI information $p$, which is a triple of an identifier, specific location $l_p$, and category ${\category_p}$ (e.g., factory). 
\end{definition}

\begin{definition}[Road network data]
A road network $\roaddata$ consists of road segments $r$. 
Each road segment includes coordinates of the start and end points, and road category $\category_r$ (e.g., highway).
\end{definition}

\begin{definition}[Meteorology data]
Meteorology data $\meteodata$ consist of distinct-level meteorological information.
Meteorological information includes meteorological measurements, such as weather and temperature. The meteorology data are time-dependent data.
\end{definition}

In this study, we aim to infer spatially fine-grained air quality in the target unmonitored city.

\smallskip
\noindent
{\bf Problem statement}.
Given target city $\targetcity$, target location $\targetlocation$ in $\targetcity$, a set $\sourcecities$ of source cities, a set of monitoring stations in $\sourcecities$, air pollutant data $\airdata$, PoI data $\poidata$, road network data $\roaddata$, and meteorology data $\meteodata$, 
we aim to infer air quality of $\targetlocation$ over time domain $T$.

\smallskip
We focus on regression for evaluating quantities of air pollutants in this paper, but our models can be used for classification for evaluating the air quality index~\cite{Cheng2018AStations}.


\section{Proposal}

We present our neural network-based architecture AIREX and training method after describing our framework and feature extraction.

\begin{figure}[t]
	\includegraphics[width=1.0\linewidth]{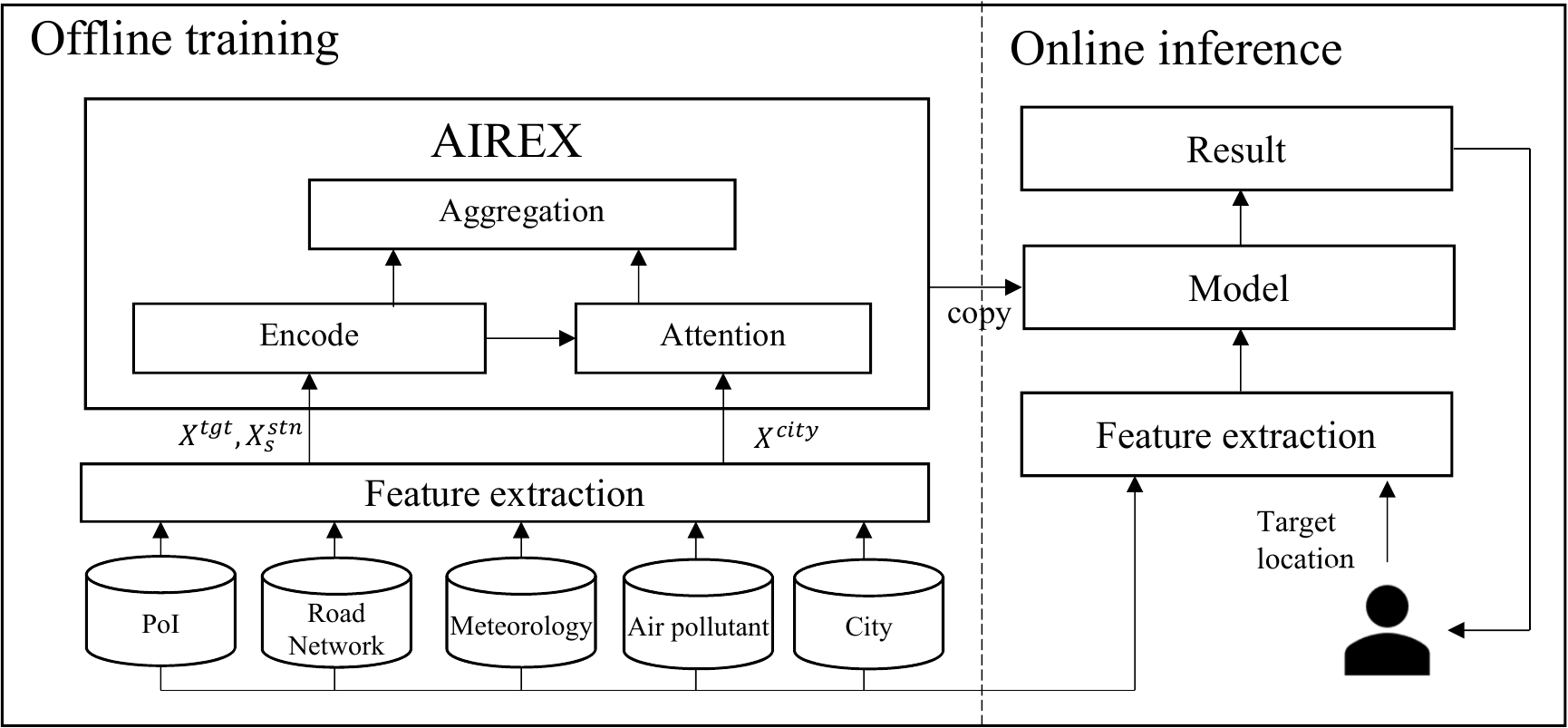}
    \caption{Our framework for air quality inference}
    \label{fig:framework}
\end{figure}

\subsection{Framework and Design Policy}
Figure~\ref{fig:framework} illustrates our framework.
This framework consists of offline training and online inference.
In the offline training, we build our inference model after extracting features, and in the online inference, we infer the air quality of the given target location by using the built model.

We describe a design policy of the offline training.
Air quality of the target location is assessed by data related to target location, monitoring station, and cities.
Thus, our framework extracts features of target location, monitoring stations, and cities, from data sources.
We leverage these features to capture the correlations of air quality between the target and source cities, and the target location and monitoring stations.


We design our inference architecture AIREX for automatically capturing the correlations and being trained in an unsupervised manner.
For this purpose, AIREX is based on the mixture-of-experts approach~\cite{Guo2018Multi-sourceExperts} and attention mechanism~\cite{bahdanau2014neural}.
The mixture-of-experts approach compute the final output by aggregating the output of multiple models (i.e., {\it experts}). 
In AIREX, each expert is a model for inferring air quality by using data of source city. 
Each source city and monitoring station does not equally contribute the air quality inference in the target city, and thus we use the attention mechanism to compute the importance of cities and monitoring stations.
AIREX can accurately infer the air quality in the target city by elegant combination of the mixture-of-experts approach and attention mechanism.
Furthermore, AIREX can be trained in an unsupervised manner by using the meta-training approach~\cite{Guo2018Multi-sourceExperts} and multi-task learning~\cite{caruana1997multitask}.
We describe the training method later.

AIREX consists of three main components: encoding, attention, and aggregation.
First, in the encode, it encodes raw input features to obtain latent features for capturing interactions between inferred values and raw input. 
Then, in the attention, AIREX computes the importance of source cities and monitoring stations for inferring air quality of the target city.
Finally,  in the aggregation, it computes output of experts for each source city by aggregating the transformed features and importance of monitoring stations, and then compute the final output by aggregating the outputs of experts and importance of cities.

\subsection{Feature extraction}
We introduce our features for assessing air quality at $\targetlocation$. 
We extract the three features, namely, the target location feature $\featuretarget$, monitoring station feature $\featurestation$, and city feature $\featurecity$.
These features comprise (1) PoI factor, (2) road network factor, (3) meteorological factor, (4) air pollutant factor, (5) station location factor, and (6) city location factor. 
We describe our three features after explaining how to extract each factor from the data.

The PoI, road network, and meteorological factors are associated with location $l$ (e.g., locations of monitoring stations and the target location). 
$l$ has its own factors that are extracted from the data within affecting region $\mathcal{L}(l)$. We set $\mathcal{L}(l)$ as a circle whose center and radius are $l$ and $1$ km, respectively. 


\smallskip
\noindent
{\bf PoI factor} $X^P_l$: 
$X^P_l$ includes the numbers of PoIs, which represents the characteristics of locations, such as the numbers of factories and public parks.
We consider a set $\Category{P}$ of PoI categories and count the number of PoIs belonging to each PoI category.
Let $X^P_l=\{x_{\category}^P(l) \}_{\category \in \Category{P}}$ denote the PoI factor for $l$. 
We compute $x_{\category}^P$ as follows:
\begin{equation}
x_{\category}^{P}(l)=|\{p \in \poidata | l_p \subset \mathcal{L}(l) \land \category_p = \category \}|.
\end{equation}

\noindent
{\bf Road network factor} $X^R_l$: 
$X^R_l$ includes the numbers of road segments, which affects local air quality, as vehicles are one of the sources of air pollutants.
We consider a set $\Category{R}$ of road categories and count the number of road segments belonging to each road category.
Let $X^R_l=\{x_{\category}^R(l) \}_{\category \in \Category{R}}$ denote the road network features extracted for $l$. 
We define $\overline{r}$ as arbitrary points between the start and end of road segment $r$.
We compute $x_{\category}^R$ as follows:
\begin{equation}
x_{\category}^{R}(l)=|\{r \in \roaddata | \overline{r} \subset \mathcal{L}(l) \land \category_r = \category \}|.
\end{equation}

\noindent
{\bf Meteorological factors} $X^M_l$:
$X^M_l$ is the sequence of meteorological measurements of $l$, such as weather and temperature, which influences the concentrations and flows of air pollutants.
The meteorological measurements have two types of values; categorical values (e.g., weather and wind direction) and numerical values (e.g., temperature and wind speed).
For categorical and numerical values, we adopt one-hot encoding and raw values, respectively.
We denote the meteorological factor at time step $t$ as $X^{Mt}_l$.

\smallskip
\noindent
These factors have demonstrated their usefulness in previous studies \cite{Xu2016WhenInference,Cheng2018AStations}.
We normalize numerical values in factors by dividing the largest values among each factor.

The monitoring and station location factors are associated with station $s$, and the city location factor is associated with city $c$.

\smallskip
\noindent
{\bf Monitoring factor} $X^A_s$: 
Quantities of air pollutants monitored by station $s$ represent the most important information for inferring air quality.
$X^A_s$ is the sequence of air pollutant quantities in $\airdata$ of station $s$.
We denote the monitoring factor at time step $t$ as $X^{At}_s$.

\noindent
{\bf Station location and city location factors} $X^C_c$ and  $X^S_s$:
The distance and direction from a location to another location are likewise important factors to measure the influence of their respective air quality levels. 
$X^S_s$ (resp. $X^C_c$) is the relative position that depecits the distance and angle from station $s$ (resp. source city $c$) to the target location $\targetlocation$ (resp. target city $\targetcity$).
\smallskip

Our features combine the above factors.
The target location feature $\featuretarget$, monitoring station feature $\featurestation$, and city feature $\featurecity$ are given as follows:
\begin{eqnarray}
\featuretarget &=& X_{\targetlocation}^{P} \cup X_{\targetlocation}^{R} \cup X_{\targetlocation}^{M}, \nonumber \\
\featurestation &=& \textit{X}_{l_{s}}^{P} \cup X_{l_{s}}^{R} \cup X_{l_{s}}^{M} \cup X_{s}^{A} \cup X_{s}^{S}, ~\mathrm{and} \nonumber \\
\featurecity &=& \cup_{c \in \sourcecities}\{X_{c}^{C}\}. \nonumber
\end{eqnarray}

Here, since the air quality changes time by time, it is preferable that all factors are time-dependent. Due to limited data sources, it is necessary to support both time-independent and time-dependent data. 

\subsection{Inference architecture}
\begin{figure}[t]
	\includegraphics[width=1.0\linewidth]{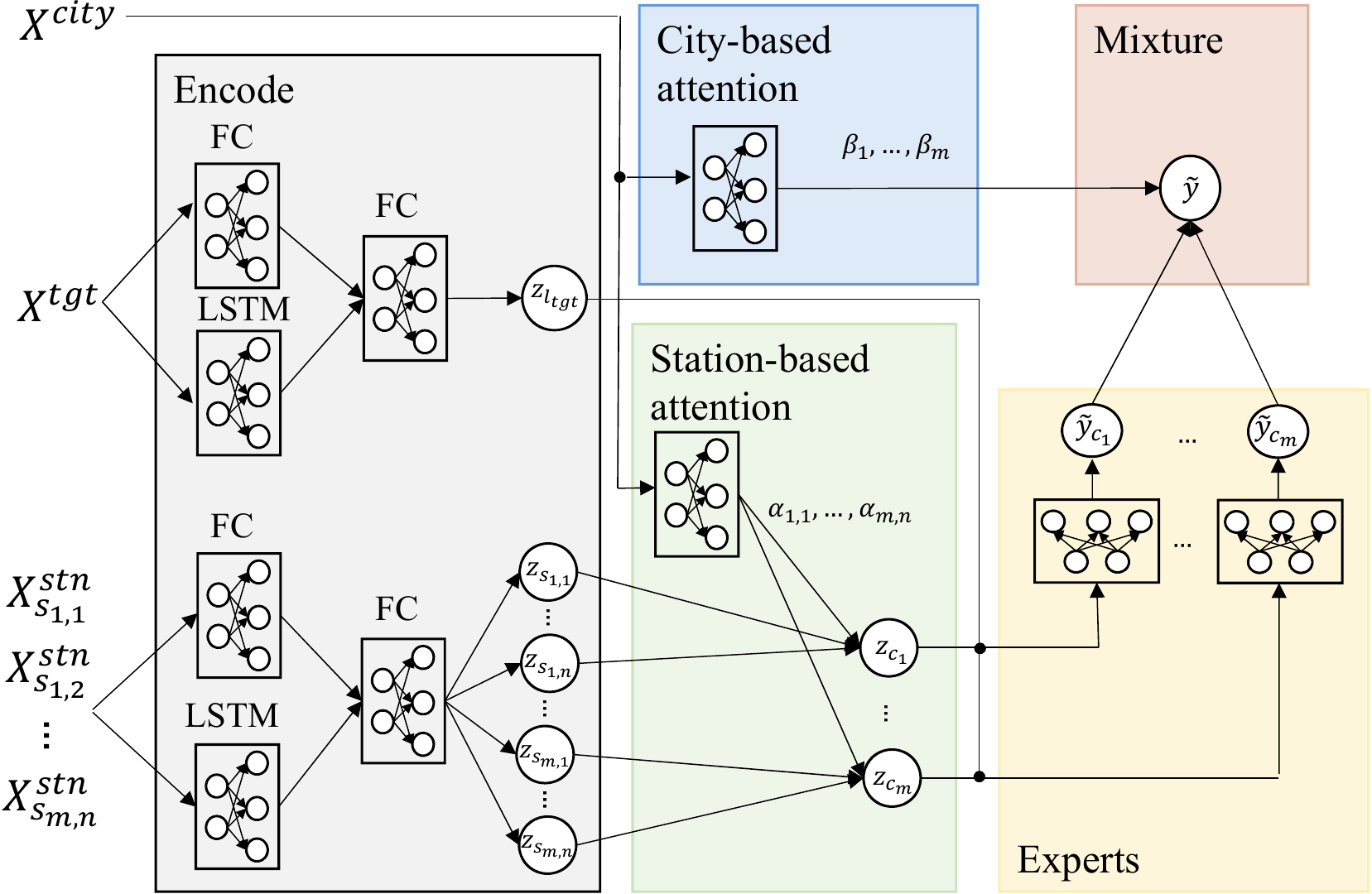}
    \caption{Neural network structure of AIREX}
    \label{fig:airex}
\end{figure}

We introduce our inference architecture AIREX. 
Figure \ref{fig:airex} shows components of AIREX.
AIREX has three input types: $\featuretarget$, $\featurestation$ for $\forall s \in S$, and $\featurecity$, and it contains five layers: encode, station-based attention, city-based attention, experts, and mixture layers.
We describe each layer in the following.

\smallskip
\noindent
{\bf Encode layer}:
The encode layer transforms $\featuretarget$ and $\featurestation$.
Each feature includes time-independent (e.g., PoI) and time-dependent (e.g., meteorology) data.
We transform time-dependent and time-independent factors by LSTM and FC, respectively~\cite{Cheng2018AStations}.
We use different models for $\featuretarget$ and $\featurestation$ because they include different factors; however, we use the same LSTM and FC for all monitoring stations to increase generalization ability.

We first explain models for time-depending factors in $\featurestation$.
$X_{l_{s}}^{M}$ and $X_{l_{s}}^{A}$ at time step $t$ are transformed into $\textit{\textbf{h}}_{s}^{t}$ as follows:
{\small
\begin{eqnarray}
\textbf{i}_{s}^{t} &\!\!\!\!\!\!=&\!\!\!\!\!\! \sigma(\textbf{W}_{ix}\textit{(X}_{l_{s}}^{Mt}\oplus\textit{X}_{l_{s}}^{At})+\textbf{W}_{ih}\textbf{h}_{s}^{t-1}+\textbf{W}_{ic}\odot\textbf{c}_{s}^{t-1}+\textbf{b}_{i}) \nonumber \\
\textbf{f}_{s}^{t} &\!\!\!\!\!\!=&\!\!\!\!\!\! \sigma(\textbf{W}_{fx}\textit{(X}_{l_{s}}^{Mt}\oplus\textit{X}_{l_{s}}^{At})+\textbf{W}_{fh}\textbf{h}_{s}^{t-1}+\textbf{W}_{fc}\odot\textbf{c}_{s}^{t-1}+\textbf{b}_{f}) \nonumber \\
\textbf{c}_{s}^{t} &\!\!\!\!\!\!=&\!\!\!\!\!\! \textbf{f}_{s}^{t}\odot\textbf{c}_{s}^{t-1}+\textbf{i}_{s}^{t}\odot tanh(\textbf{W}_{cx}\textit{(X}_{l_{s}}^{Mt}\oplus\textit{X}_{l_{s}}^{At})+\textbf{W}_{ch}\textbf{h}_{s}^{t-1}+\textbf{b}_{c}) \nonumber \\
\textbf{o}_{s}^{t} &\!\!\!\!\!\!=&\!\!\!\!\!\! \sigma(\textbf{W}_{ox}\textit{(X}_{l_{s}}^{Mt}\oplus\textit{X}_{l_{s}}^{At})+\textbf{W}_{oh}\textbf{h}_{s}^{t-1}+\textbf{W}_{oc}\odot\textbf{c}_{s}^{t}+\textbf{b}_{o}) \nonumber \\
\textbf{h}_{s}^{t} &\!\!\!\!\!\!=&\!\!\!\!\!\! \textbf{o}_{s}^{t}\odot tanh(\textbf{c}_{s}^{t}) \nonumber
\end{eqnarray}
}
\noindent 
where, $\textit{\textbf{W}}$ is weight matrix,  $\textit{\textbf{b}}$ is bias vector, and $\odot$ indicates Hadamard product.
$\textit{\textbf{i}}$, $\textit{\textbf{f}}$, $\textit{\textbf{o}}$, $\textit{\textbf{c}}$, and $\textit{\textbf{h}}$ are input gate, forget gate, output gate, memory cell, and final states of hidden layer, respectively.

Next, we describe models for time-independent factors.
$X_{l_{s}}^{P}$, $X_{l_{s}}^{R}$, and $X_{l_{s}}^{S}$ in $\featurestation$ are translated into embedding $\textbf{z}_{s}^{(n)}$ as follows:
{\small
\begin{eqnarray}
  \textbf{z}_{s}^{(i)}=
  \begin{cases}
  ReLU(\textbf{W}_{s}^{(i)}(X_{l_s}^{P}\oplus X_{l_s}^{R} \oplus X_{l_{s}}^{S})+\textbf{b}_{s}^{(i)}), i=1 \\
  ReLU(\textbf{W}_{s}^{(i)}\textbf{z}_{s}^{ni1}+\textbf{b}_{s}^{(i)}), 1 < i \leq L \\
  \end{cases}
  \nonumber
\end{eqnarray}
}
\noindent
where $L$ denotes the number of hidden layers.



$\featuretarget$ is transformed in the same way as $\featurestation$.
The difference is the input factors.

Finally, the transformed features generated by the LSTM and FC are concatenated to input another FC to obtain the features $\textbf{z}_{*}^{(n')}$ as follows:
{\small
\begin{eqnarray}
  \textbf{z}_{*}^{(i')}=
  \begin{cases}
  ReLU(\textbf{W}_{*'}^{(i')}(\textbf{z}_{*}^{L}\oplus\textbf{h}_{*}^{t})+\textbf{b}_{*'}^{(i')}), i'=L+1 \\
  ReLU(\textbf{W}_{*'}^{(i')}\textbf{z}_{*}^{i'-1}+\textbf{b}_{*'}^{(i')}), i' \in [L+2,L+L'] \\
  \end{cases}
  \nonumber
\end{eqnarray}
}
\noindent
where $*$ indicates either $\targetlocation$ or $s$ and $L'$ denotes the number of hidden layers.


\smallskip
\noindent
{\bf City-based Attention layer}:
Not all source cites contribute equally to inference in the target city. 
AIREX automatically captures the importance of different city data by employing the attention mechanism.
The city-based attention layer computes {\it city-attention factor} which represents the weights of influences of source cities to air quality in the target city.
The city-attention factor $\beta_{c_{k}}$ of source city $c_k$ is computed as follows: 
{\small
\begin{eqnarray}
\textbf{z}_{\oplus k}^{(L+L')} &\!\!\!\!\!\!=&\!\!\!\!\!\! \textbf{z}_{s_{k,1}}^{(L+L')}\oplus\cdots\oplus\textbf{z}_{s_{k,n}}^{(L+L')} \nonumber \\
\beta'_{c_{k}} &\!\!\!\!\!\!=&\!\!\!\!\!\! \textbf{w}_{\beta}^{\intercal}ReLU(\textbf{W}_{\beta}(\textbf{z}_{\targetlocation}^{(L+L')}\oplus\textbf{z}_{\oplus k}^{(L+L')}\oplus X_{c_{k}}^{C})+\textbf{b}_{\beta})+b_{\beta} \nonumber \\
\beta_{c_{k}} &\!\!\!\!\!\!=&\!\!\!\!\!\! \frac{exp(\beta'_{c_{k}})}{\Sigma_{c\in \sourcecities}{exp(\beta'_{c})}} \nonumber
\end{eqnarray}
}
\noindent
{\bf Station-based Attention layer}:
Each monitoring station has a different impact to the target location, as distances and angles between each monitoring station and target location are different as well as similarity of their features. 
In the station-based attention layer, we compute {\it station-affect factor}, which is a weight of influence of monitoring stations on the air quality of the target location.
The station-affect factor $\alpha_{k, i}$ for stations $s_i$ in source city $c_k$ is calculated by the following equation:

{\small
\begin{eqnarray}
\alpha'_{k,i} &=& \textbf{w}_{\alpha}^{\intercal}ReLU(\textbf{W}_{\alpha}(\textbf{z}_{\targetlocation}^{(L+L')}\oplus\textbf{z}_{s_{k,i}}^{(L+L')})+\textbf{b}_{\alpha})+b_{\alpha} \nonumber \\
\alpha_{k,i} &=& \frac{exp(\alpha'_{k,i})}{\Sigma_{s_{i}\in S_{k}}{exp(\alpha'_{k,i})}} \nonumber
\end{eqnarray}
}
We then compute embedding $\textit{\textbf{z}}_{c_k}$ of source city with station affect-factors as follows:
\begin{eqnarray}
\textit{\textbf{z}}_{c_k}=\sum_{s_{i}\in S_{k}}{\alpha_{k,i}\textbf{z}_{s_{k,i}}^{(L+L')}}. \nonumber
\end{eqnarray}
\noindent
$\textit{\textbf{z}}_{c_k}$ represents how much is influence air quality of source city $c_k$ to the target location.

\smallskip
\noindent
{\bf Experts layer}:
The experts layer computes an inferred value on each source city.
Inferred value $\tilde{y}_{c_{k}}$ of $c_k$ is computed by the following equation:
{\small
\begin{eqnarray}
\tilde{y}_{c_{k}}=\textbf{w}_{k}^{\intercal}ReLU(\textbf{W}_{k}(\textbf{z}_{\targetlocation}^{(L+L')}\oplus\textit{\textbf{z}}_{c_k}))+\textbf{b}_{k})+b_{k}. \nonumber
\end{eqnarray}
}
This equation represents an expert model. We use this simple model for all cities to eliminate the the impact of performance of experts to the final output in this paper. 

\smallskip
\noindent
{\bf Mixture layer}:
We obtain the inferred value by summing outputs of experts weighted by city attention factors as follows:
{\small
\begin{eqnarray}
 \tilde{y}=\sum_{c_{k}\in \sourcecities}{\beta_{c_{k}}\tilde{y}_{c_{k}}}. \nonumber 
\end{eqnarray}
}

\subsection{Training method}
One of the major challenges of our study is the training of AIREX because we cannot directly train our model due to missing air quality data of the target city. 
We develop a training method in an unsupervised manner. 
We describe our approach and loss function in the training phase.

\smallskip
\noindent
{\bf Overall idea}: 
We employ a meta-training approach~\cite{Guo2018Multi-sourceExperts}, which supports to learn the differences between individual features and cities in an unsupervised setting.
Given a set of source cities, the meat-training approach regards a single source city as a {\it temporal} target city, and then trains models using the pair of temporal target and other source cities.
The temporal target and other source cities are referred to as the {\it meta-target} $c_t$ and {\it meta-sources} $c_i \in C_s$, respectively.
We obtain $|\sourcecities|$ training pairs of meta-target and meta-sources.

We use a multi-task learning method with a shared encoder.
We design loss functions for accurately inferring air quality and capturing the difference between source and target cities.

\smallskip
\noindent
{\bf Loss functions}: 
The main objective of our training is that the final outputs are closer to the actual value. Since we have multiple experts, we additionally train them. 
It is not sufficient to evaluate the difference between outputs and true values because we must capture the correlations between the source and target cities. 
Since we do not have air quality data in the target city, we must indirectly learn the correlations.
For this purpose, we use a loss for minimizing the difference between the transformed features of cities.
We note that the true values in training phase are air quality of the meta-targets instead of the actual target location.

The loss $\losspred$ is the main loss function for evaluating the inference accuracy.
$\losspred$ is computed by the mean squared error (MSE) between the final output $\tilde{y}$ and true value $y$ as follows:
{\small
\begin{eqnarray}
\losspred=\frac{1}{|\mathcal{T}|}\sum_{x \in \mathcal{T}}{(\tilde{y}(x)-y(x))^{2}} \nonumber
\end{eqnarray}
}
\noindent
where $\mathcal{T}$ denotes the set of training pairs.

The loss $\lossmean$ is one for evaluating the inference accuracy of an individual expert.
$\lossmean$ is computed by MSE between $\tilde{y}_{c_{k}}$ for source city $c_k$ of outputs of experts and $y$.
{\small
\begin{eqnarray}
\lossmean=\frac{1}{|C_s|}\sum_{c_{i}\in C_s}\left( \frac{1}{|\mathcal{T}|}\sum_{x \in \mathcal{T}}{(\tilde{y}_{c_{i}}(x)-y(x))^{2}}\right). \nonumber
\end{eqnarray} 
}

The loss $\lossadv$ is for evaluating the difference of cities. 
It is computed based on maximum mean discrepancy (MMD)~\cite{Gretton2012ATest} as the adversary to minimize the divergence between the marginal distribution of target and source cities. MMD is known as effective distance metric measures for evaluating the discrepancy between two distributions explicitly in a non-parametric manner.

{\small 
\begin{eqnarray}
\lossadv &=& {\mathit MMD}^{2}(\textbf{z}_{\cup c_1}\cup\cdots\cup\textbf{z}_{\cup c_{|C_s|}},\textbf{z}_{\cup c_t}), \nonumber \\
\textbf{z}_{\cup c_i} &=& \cup_{s \in S_i} \textbf{z}_{s}^{(L+L')}, \nonumber \\
{\mathit MMD}(\mathcal{X},\mathcal{X'}) &=& \bigg\rvert\bigg\rvert\frac{1}{|\mathcal{X}|}\sum_{\textbf{x}\in\mathcal{X}}{\phi(\textbf{x})}-\frac{1}{|\mathcal{X'}|}\sum_{\textbf{x}'\in\mathcal{X'}}{\phi(\textbf{x}')} \bigg\lvert\bigg\lvert_{\mathcal{H}}, \nonumber 
\end{eqnarray}
}
In MMD computation, $\mathcal{H}$ indicates the reproducing kernel Hilbert space (RKHS) and $\phi$ is the mapping function to RKHS space.
In our method, we compute the MMD score by the kernel method~\cite{Bousmalis2016DomainNetworks}.
The kernel method computes the MMD score as follows:
{\small
\begin{eqnarray}
{\mathit MMD}(\mathcal{X},\mathcal{X'}) &=&  \frac{1}{|\mathcal{X}|(|\mathcal{X}|-1)}\sum_{\textbf{x},\textbf{x}' \in \mathcal{X}, \textbf{x} \neq \textbf{x}'}{\mathcal{K}(\textbf{x}, \textbf{x}')} \nonumber \\
&+& \frac{1}{|\mathcal{X'}|(|\mathcal{X'}|-1)}\sum_{\textbf{x},\textbf{x}' \in \mathcal{X'}, \textbf{x} \neq \textbf{x}'}{\mathcal{K}(\textbf{x}, \textbf{x}')} \nonumber \\
&-& \frac{2}{|\mathcal{X}||\mathcal{X'}|}\sum_{\textbf{x} \in \mathcal{X}} \sum_{\textbf{x}' \in \mathcal{X}'}{\mathcal{K}(\textbf{x},\textbf{x}')} \nonumber
\end{eqnarray}
}
where, $\mathcal{K}$ indicates a Gaussian karnel function: $\mathcal{K}(\textbf{x},\textbf{x}') = exp(-\frac{1}{2\sigma^{2}}||\textbf{x}-\textbf{x}'||^{2})$.

We further use regularization of $\beta$ to avoid overfitting.
The regularization computes the entropy of $\beta$ and the sum of them.
{\small
\begin{eqnarray}
\mathcal{R} = \sum_{c_{i}\in C_s}{\beta_{c_{i}} \log \beta_{c_{i}}}
\end{eqnarray}
}
The total loss function to be minimized in our training phase is defined as follows:
\begin{eqnarray}
\mathcal{L}=\lambda\cdot\losspred+(1-\lambda)\cdot\lossmean+\gamma\cdot\lossadv+\zeta\cdot\mathcal{R} \nonumber
\end{eqnarray}
where $\lambda$, $\gamma$, and $\zeta$ are hyper parameters.




\section{Experiments}
In this section, we evaluate the inference accuracy of AIREX compared with the state-of-the-art methods.
We aim to validate that AIREX can accurately infer air quality in unmonitored cities and other methods cannot\footnote{Please see a supplementary file for detail implementation, data statistics, and additional results.}.

\subsection{Experimental settings}
\noindent
{\bf Dataset}:
We use data of 20 cities in China spanning four months from June 1st 2014/6/1 to September 30th 2014.
We collect air quality data, road network, PoI, and meteorology data as follows.
Air quality data is provided as open data by Microsoft\footnote{\url{www.microsoft.com/en-us/research/project/urban-computing/}}.
We focus on inferring PM$_{2.5}$.
We collect PoI data from Foursquare\footnote{\url{developer.foursquare.com}} and categorize them into ten categories according to the official categories provided by Foursquare. 
For road network data, we use OpenStreetMap\footnote{\url{www.openstreetmap.org/}}, and roads are categorised into three types; highway, trunk, and other.
For meteorology data, we use weather, temperature, air pressure, humidity, wind speed, and wind direction, which is also provided by Microsoft.
Air quality and meteorology data are sampled every hour.

\smallskip
\noindent
{\bf Evaluation}:
We select four cities as target cities; Beijing, Tianjin, Shinzhen, and Guangzhou.
Beijing and Tianjin are cities in the northern area of China, whereas Shinzhen and Guangzhou are in the south.
We randomly select five monitoring stations from each city for training and test data.
The ratio of training and test data is $|\sourcecities|$ to one.

As evaluation metrics of inference accuracy, we use the root mean squared error (RMSE) for PM$_{2.5}$, which is a standard metric \cite{Xu2016WhenInference,Cheng2018AStations}. We run three times for training by changing monitoring stations. 


\noindent
{\bf Compared methods and hyper parameters}:
We compare AIREX with three approaches:
(a) k nearest neighbors (KNN): This method selects the $k$ monitoring stations closest to the target location, and compute the average air pollutant quantities from these stations as result. 
We set $k$ to be three in our experiments.
(b) Feedforward neural networks (FNN): This method uses a simple neural network model, whose inputs are $\featuretarget$ and $\featurestation$ for all stations.
In our experiments, the model consists of three layers with 200 units.
For sequential features, we only use their values at the same time step of the inferred air quality.
(c) ADAIN: This method represents the state-of-the-art neural network model for inferring air quality \cite{Cheng2018AStations}. We use two cases of source cities: ADAIN5 and ADAIN19, whose source cities are the five cities closest to the target city and all source cities, respectively.

In parameter settings of AIREX and ADAIN, we follow the setting in experiments of ADAIN~\cite{Cheng2018AStations}. 
We construct a single basic FC layer ($L=1$) with 100 neurons and two LSTM layers
with 300 memory cells per layer. We then build two layers of the high-level FC network ($L'=2$) with 200 neurons per layer. 
The time-dependent data is input in 24 time steps (i.e., one day).
The number of epochs, the batch size, learning rate are selected from [100, 200, 300], [32, 64, 128, 256, 512], and [0.005, 0.01], respectively, by grid search.
In our model, $\lambda$,  $\gamma$, and $\zeta$ in AIREX are 0.5, 1.0, and 1.0, respectively.
Further detail is provided in our codes.

\subsection{Experimental results}
\begin{figure}[t]
    \centering
	\includegraphics[width=0.9\linewidth]{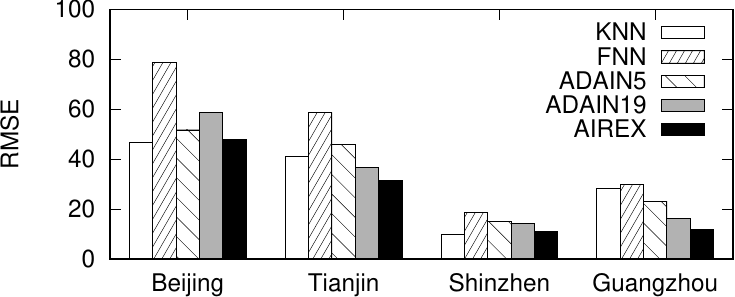}
    \caption{An overview of accuracy}
    \label{fig:result}
\end{figure}

Figure \ref{fig:result} shows the inference accuracy for each method.
AIREX achieves the best accuracy in Tianjin and Guanzhou and the second best in Beijing and Shinzhen.
Since AIREX learns the difference between target and source cities, it can accurately infer air quality without air quality data in the target city. 
KNN achieves the best accuracy in Beijing and Shinzhen, as these are monitoring stations that very close to the target location, whereas KNN  fails the accurate inference when there are no monitoring stations close to the target location like Tianjin and Guangzhou.
ADAIN and FNN do not perform well in all target cities.
In particular, although ADAIN is the state-of-the-art method for inferring air quality, it does not perform well when the source and target cities are different.

We further investigate the difference between AIREX and ADAIN, as ADAIN may perform well if we use optimal source cities.
Table \ref{tab:beijing_dist} shows the accuracy of AIREX and ADAIN in Beijing and Guangzhou as target cities (see appendix for Tianjin and Shenzhen).
In ADAIN, we use each city as the source city in addition to ADAIN5 and ADAIN19.
In Beijing, ADAIN accurately infers the air quality when its source city is Beijing (i.e., target and source cities are the same).
However, the accuracy of ADAIN significantly decreases when ADAIN uses different cities even when the source cities are close to Beijing. 
In Guangzhou, AIREX achieves better performance than ADAIN even when ADAIN uses Guangzhou as the source city. This result indicates that the use of multiple cities increases the inference accuracy if we can capture the correlations of air quality between cities.
From these results, we can confirm that our mixture-of-experts approach combined with attention mechanisms performs well for accurately inferring the air quality in unmonitored cities without selecting source cities.

\begin{table}[t]
\caption{AIREX vs ADAIN in different source cities. City names indicate the result obtained by ADAIN, where training data is the city. A distance of zero kilometers indicates that the target and source cities are the same.}
\label{tab:beijing_dist}
\centering
{\small
\begin{tabular}{|ll|r|r|r|r|}
\hline
\multicolumn{2}{|c|}{\multirow{2}{*}{Method}} & \multicolumn{2}{c|}{Beijing} & \multicolumn{2}{c|}{Guangzhou} \\ \cline{3-6}
& & \multicolumn{1}{|c|}{RMSE} & \multicolumn{1}{c|}{Dist. {[}km{]}} & \multicolumn{1}{c|}{RMSE} & \multicolumn{1}{c|}{Dist. {[}km{]}} \\ \hline
\multicolumn{2}{|c|}{AIREX} &47.88& --- &11.90& ---\\\hline
\multirow{21}*{\rotatebox[origin=c]{90}{ADAIN}}&5 NN cities &51.70&--- &23.05& ---\\
&19 cities&58.83& --- &16.34& ---\\
&Beijing &30.49& 0  &78.15& 1883.5\\
&Langfang & 48.58 & 47.1  &61.57& 1847.9 \\ 
&Tianjin & 52.05 & 113.8  &43.01& 1807.9\\ 
&Baoding & 68.46 & 140.3  &52.01& 1758.1\\ 
&Tangshan& 60.53 & 154.9  &54.85& 1887.7\\ 
&Zhangjiakou & 81.79 & 160.9  &24.0& 1961.8\\ 
&Chengde & 74.67 & 176.0  &27.18& 2024.9\\ 
&Cangzhou & 59.57 & 181.5  &47.22& 1716.5\\ 
&Hengshui & 66.90 & 248.8  &52.68& 1635.9\\ 
&Shijiazhuang & 55.80 & 263.8  &87.56& 1657.7\\ 
&Qinhuangdao& 69.13 & 273.0  &43.79& 1956.8\\ 
&Zibo & 69.62 & 372.1  &41.98& 1585.2\\ 
&Shantou & 85.63 & 1,835.3  &25.95& 350.6\\ 
&Huizhou & 84.25 & 1,871.4  &22.30& 118.0\\ 
&Guangzhou & 76.62 & 1,883.5  &19.70& 0\\ 
&Dongguan & 80.07 & 1,888.8  &20.29& 51.4\\ 
&Foshan & 76.56 & 1,897.4  &21.87& 18.9\\ 
&Shenzhen & 86.14 & 1,937.7  &23.64& 104.1\\ 
&Jiangmen & 82.32 & 1,946.5  &22.75& 63.8\\ 
&Hong Kong & 86.99 & 1,953.3  &26.46& 118.8\\ \hline
\end{tabular}
}

\end{table}
\section{Related Work}

We review neural network-based approaches for spatially fine-grained air quality inference.
Numerous methods have been proposed \cite{Shad2009PredictingGIS,Hasenfratz2014PushingMaps,Xu2016WhenInference}, such that employ linear regression, matrix factorization and neural networks.
For example, Zheng et al.~\cite{Zheng2013U-Air:Data} proposed U-air, which is a neural network-based classifier model that captures both spatial and temporal correlations.
Hu et al.~\cite{Hu2018Real-timeOptimization} proposed an architecture that employs deep reinforcement learning for optimizing air quality sensing systems.
Zhong et al.~\cite{zhong2020airrl} proposed AirRL, which consists of station selector that distills  monitoring stations using reinforcement learning.
Cheng et al.~\cite{Cheng2018AStations} proposed ADAIN, which employs the attention mechanism to assign weights to station-oriented features. We used ADAIN to design encode and station-based attention layers.
To the best of our knowledge, we first employ a mixture-of-experts approach for air quality inference.

None of them addresses the problem of air quality inference in unmonitored cities.
In contrast to these studies, our method assigns weights to each city automatically, without selecting monitoring stations.
\section{Conclusion}
We addressed a new problem that infers air quality information in unmonitored cities.
For the problem, we proposed AIREX, which can accurately infer air quality in unmonitored cities.
Experimental studies using real data showed that AIREX outperforms the state-of-the-art methods.

As our future works, we address air quality inference in different countries, in particular, countries across sea, and support environments that each city has different data sources.



\bibliographystyle{named}
\bibliography{airquality}

\end{document}


\maketitle

\setcounter{table}{2}
\section{Dataset statistics}

Table~\ref{tb:dataset} shows data statistics of each city.
Each city has a different number of stations and PM$_{2.5}$ value.
We can see that statistics of PM$_{2.5}$ values are different among cities.
In particular, air quality in cities placed in northern and southern areas have huge gaps.

\begin{table}[!b]
\centering
 {\scriptsize
\caption{Data statistics of each city}
\label{tb:dataset}
\begin{tabular}{cccccc}
\hline
City & Area & \# of stations & Range & Average & Variance\\ \hline
BeiJing & North & 36 & [2.00, 389.00] & 71.84 & 54.72\\ 
Tianjin & North & 27 & [1.00, 645.00] & 66.50 & 40.31\\ 
Shijianzhuan & North & 24 & [1.00, 816.00] & 94.09 & 60.58\\ 
Tangshan & North & 18 & [3.00, 617.00] & 89.86 & 54.45\\ 
Qinhuangdao & North & 9 & [1.00, 525.00] & 60.93 & 49.64\\ 
Baoding & North & 27 & [1.00, 705.00] & 93.91 & 48.17\\ 
Zhangjiakou & North & 19 & [1.00, 628.00] & 36.38 & 31.57\\ 
Chengde & North & 14 & [1.00, 462.00] & 50.72 & 38.55\\ 
Cangzhou & North & 16 & [1.00, 359.00] & 72.30 & 35.66\\ 
Langfang & North & 12 & [1.00, 496.00] & 84.60 & 55.57\\ 
Hengshui & North & 11 & [1.00, 1273.00] & 91.32 & 40.72\\ 
Zibo & North & 12 & [8.00, 275.00] & 75.63 & 26.50\\ 
Shenzhen & South & 11 & [1.00, 122.00] & 18.52 & 14.04\\ 
Guangzhou & South & 42 & [1.00, 282.00] & 34.97 & 20.44\\ 
Hong Kong & South & 15 & [1.00, 118.00] & 15.92 & 12.16\\ 
Dongguan & South & 5 & [6.00, 120.00] & 30.06 & 15.52\\ 
Foshan & South & 8 & [2.00, 161.00] & 28.49 & 20.81\\ 
Huizhou & South & 7 & [1.00, 147.00] & 23.25 & 14.87\\ 
Jiangmen &South & 7 & [1.00, 400.00] & 23.48 & 18.16\\ 
Shantou & South & 6 & [1.00, 105.00] & 21.79 & 14.73\\ 
\hline
\end{tabular}
 }
\end{table}

\section{Training and inference time}
Experiments were performed on a Linux server with 64GB of memory, an Intel(R) Xeon(R) CPU E5-2640 v4 @2.10GHz processor, and NVIDIA Tesla K40 of GPU. 
Our model takes 0.7 seconds for inference and 1.48 hours as training for one epoch.

\section{AIREX vs ADAIN}

Table 1 in the main body shows the inference accuracy in Beijing and Guanzhou.
Table~\ref{tab:tianjin_dist} shows the inference accuracy in Tianjin and Shenzhen.
These results are similar tendency to the result of Beijing and Guanzhou.

\begin{table}[t]
\caption{AIREX vs ADAIN in different source cities. City names indicate the result obtained by ADAIN, where training data is the city. A distance of zero kilometers indicates that the target and source cities are the same.}
\label{tab:tianjin_dist}
\centering
{\scriptsize
\begin{tabular}{|ll|r|r|r|r|}
\hline
\multicolumn{2}{|c|}{\multirow{2}{*}{Method}} & \multicolumn{2}{c|}{Tianjin} & \multicolumn{2}{c|}{Shenzhen} \\ \cline{3-6}
& & \multicolumn{1}{|c|}{RMSE} & \multicolumn{1}{c|}{Dist. {[}km{]}} & \multicolumn{1}{c|}{RMSE} & \multicolumn{1}{c|}{Dist. {[}km{]}} \\ \hline
\multicolumn{2}{|c|}{AIREX} &31.60& --- &10.90& ---\\\hline
\multirow{21}*{\rotatebox[origin=c]{90}{ADAIN}}&5 NN cities &45.86&--- &14.99 & ---\\
&19 cities&36.76& --- &14.35& ---\\
&Beijing &50.69& 113.8  & 70.33 & 1,937.7 \\
&Langfang & 45.44 & 67.3  & 58.88& 1,900.6 \\ 
&Tianjin & 34.76 & 0  &46.25&1,858.1 \\ 
&Baoding & 54.42 & 152.2  &82.30& 1,815.5\\ 
&Tangshan& 41.93 & 104.0  &70.65& 1,934.4 \\ 
&Zhangjiakou & 57.94 & 272.2  &21.60& 2,022.3\\ 
&Chengde & 52.21 & 217.4  & 38.26 & 2,074.0\\ 
&Cangzhou & 47.87 & 92.1  & 58.91 & 1,767.4\\ 
&Hengshui & 50.49 & 200.4  & 64.77& 1,691.6 \\ 
&Shijiazhuang & 60.10 & 261.1  & 92.01& 1,718.8 \\ 
&Qinhuangdao& 49.78 & 227.0  & 28.14 & 1,998.2 \\ 
&Zibo & 50.13  &  263.0 & 69.35 & 1,628.0\\ 
&Shantou & 65.65  & 1,744.8  & 18.19& 283.7 \\ 
&Huizhou & 63.85 & 1,790.5  & 12.61 & 72.9 \\ 
&Guangzhou & 58.26 & 1,807.9  & 22.75 & 104.1 \\ 
&Dongguan & 58.50  & 1,810.8  & 16.86& 61.5 \\ 
&Foshan & 57.06 & 1,822.4  & 18.88 & 109.8\\ 
&Shenzhen & 66.61  & 1,858.1  & 10.49 & 0 \\ 
&Jiangmen &66.15  & 1,871.3  & 17.02&100.5 \\ 
&Hong Kong & 68.98  & 1,873.3 & 13.14& 17.1\\ \hline
\end{tabular}
}
\end{table}
